\documentclass[runningheads]{llncs}
\usepackage[T1]{fontenc}
\usepackage{multirow}
\usepackage{amsmath,graphicx,booktabs,amssymb,stmaryrd}
\usepackage{wrapfig}
\usepackage{subcaption}
\usepackage{caption}
\usepackage{color}
\usepackage{bbm}
\usepackage{booktabs}
\usepackage{bbding}
\usepackage[pagebackref=true,breaklinks=false,colorlinks,bookmarks=false]{hyperref}
\usepackage{algorithm}
\usepackage{bm}
\usepackage{algpseudocode}

\usepackage{marvosym}

\usepackage[dvipsnames]{xcolor}
\definecolor{222}{HTML}{00acd0}
\definecolor{227}{HTML}{E52B50}

\begin{document}
\title{\raisebox{-1.5ex}{\includegraphics[height=2em]{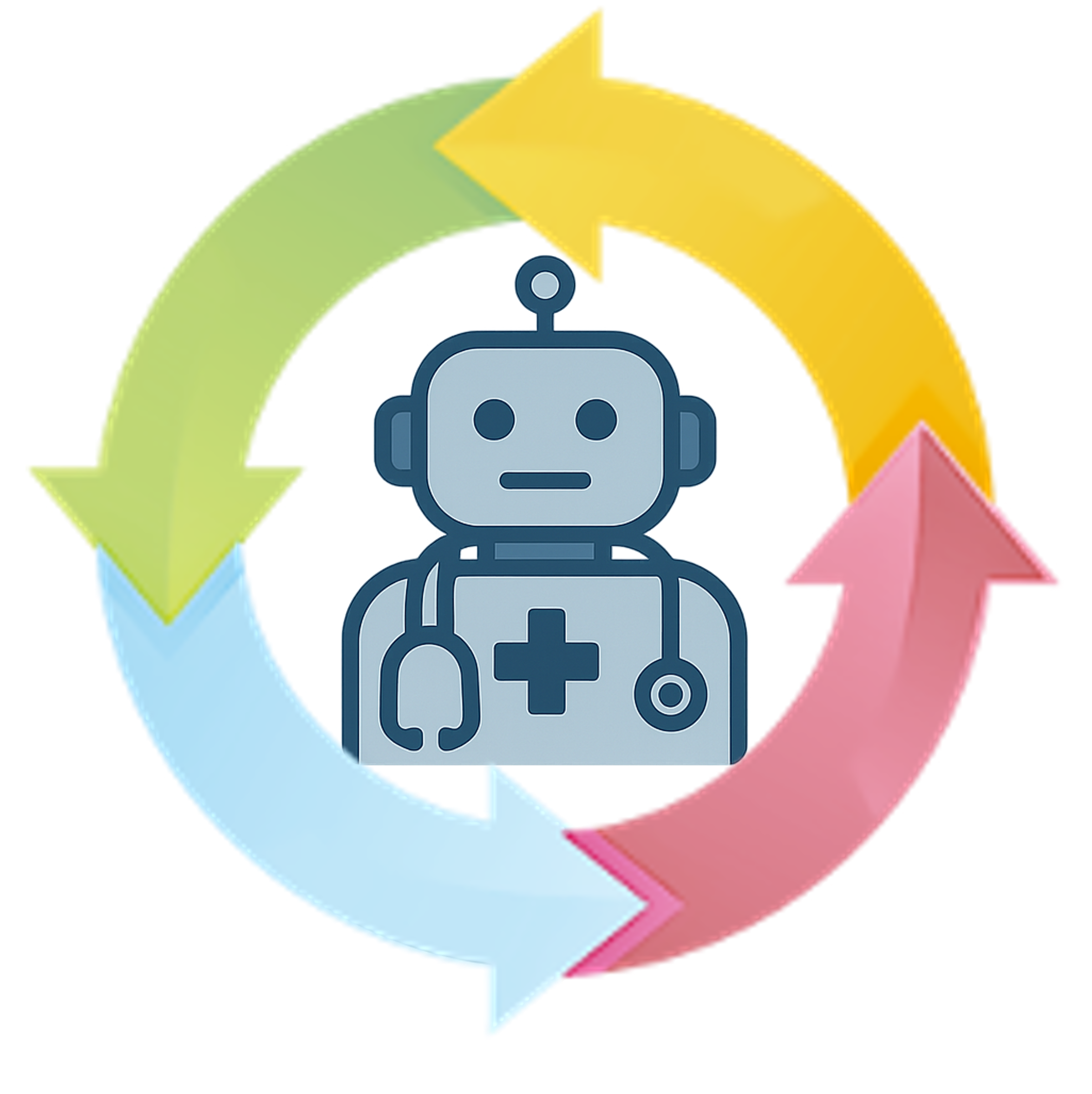}} Med-Evo: Test-time Self-evolution for Medical Multimodal Large Language Models}


\titlerunning{Med-Evo: Test-time Self-evolution for medical MLLMs}
%

\author{Dunyuan Xu\inst{1}$^{*}$ 
\and Xikai Yang\inst{1}$^{*,\dag}$ 
\and Juzheng Miao\inst{1}
\and Yaoqian Li\inst{1}
\and Jinpeng Li\inst{1}$^{\textrm{\Letter}}$ 
\and Pheng-Ann Heng\inst{1,2}}

\authorrunning{D. Xu et al.}

\institute{Department of Computer Science and Engineering, The Chinese University of Hong
Kong, Hong Kong, China \\
\email{jpli21@cse.cuhk.edu.hk}\and
Institute of Medical Intelligence and XR, The Chinese University of Hong Kong,
Hong Kong, China}
  
\maketitle              
\begin{abstract}
Medical Multimodal Large Language Models (MLLMs) have demonstrated remarkable capabilities across diverse healthcare tasks. 
However, current post-training strategies, such as supervised fine-tuning and reinforcement learning, heavily depend on substantial annotated data while overlooking the potential of unlabeled test data for model enhancement.
This limitation becomes particularly pronounced in medical domains, where acquiring extensive labeled medical data is difficult due to the strict data sensitivity and annotation complexity.
Moreover, leveraging test data poses challenges in generating reliable supervision signals from unlabeled samples and maintaining stable self-evolution.
To address these limitations, we propose Med-Evo, the first self-evolution framework for medical MLLMs that utilizes label-free reinforcement learning to promote model performance without requiring additional labeled data.
Our framework introduces two key innovations:
$1)$ Feature-driven Pseudo Labeling (FPL) that identifies semantic centroids from all heterogeneous candidate responses to select pseudo labels in each rollout, and
$2)$ Hard-Soft Reward (HSR) that combines exact match with token-level assessment and semantic similarity to provide hierarchical reward. 
Experiments on three medical VQA benchmarks and two base MLLMs show clear advantages of our approach over SOTA methods, with significant improvements of 10.43\% accuracy and 4.68\% recall on the SLAKE dataset using Qwen2.5-VL, showing the effectiveness of our method.
\footnote[1]{$^{*}$Equal contribution, $^{\dag}$Project lead}


\end{abstract}

\section{Introduction}
\begin{figure}[t]
\centering
\includegraphics[width=\linewidth]{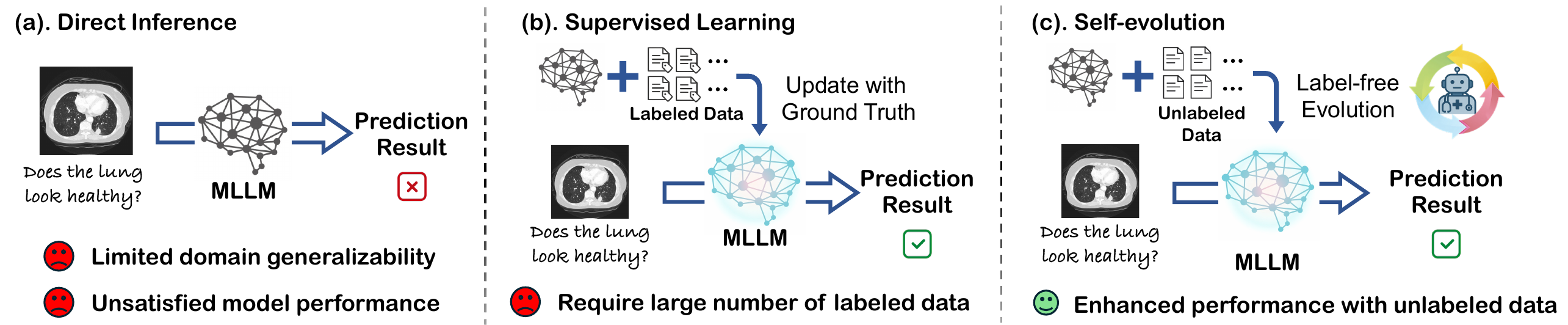}
\caption{Overview of medical MLLM adaptation strategies. (a) Direct inference leads to poor generalization on down-stream tasks; (b) Supervised learning requires extensive labeled data; (c) Our test-time self-evolution strategy using unlabeled data.}
\label{Fig:teaser}
\end{figure}

Medical Multimodal Large Language Models (MLLMs) have achieved remarkable progress due to their superior capabilities across various tasks \cite{ma2025medla} and potential to alleviate clinical workload \cite{zhang2024potential}.
However, as illustrated in Fig. \ref{Fig:teaser}(a-b), existing approaches predominantly focus on leveraging only training data for MLLM improvement through supervised fine-tuning \cite{he2024pefomed,xu2025perceive,wang2025hicur} or reinforcement learning \cite{zhi2025medgr,liu2025infimed,zhu2025toward}, while neglecting the opportunity for optimizing MLLMs by using test data. 
This limitation becomes particularly pronounced in medical domains, where the strict data sensitivity \cite{sysoykova2025federated,sood2025medical} and annotation complexity \cite{goel2023llms,wac2025capturing} make it challenging to collect extensive labeled data \cite{tajbakhsh2021guest}.
Moreover, the significant variations across diverse clinical scenarios demand MLLMs capable of adaptive and continuous improvement \cite{xu2024towards}, highlighting the need for self-evolving framework that can use unlabeled cases for ongoing performance enhancement.

While recent works have begun exploring Test-time Training (TTT) for MLLMs, leveraging test data for medical MLLM improvement still faces two key challenges. 
The first challenge lies in generating reliable pseudo labels. 
Existing methods address this by generating multiple candidate answers (i.e., rollouts) for each test case and applying majority voting \cite{abdullahi2024learning,wei2025first,zuo2025ttrl,hosseini2025cg,zhang2025aqa} to select the most frequent one. 
However, this approach fails in medical VQA scenarios where complex medical reasoning produces heterogeneous responses \cite{rawat2024diversitymedqa}, making it difficult to identify a dominant answer for stable pseudo label selection. 
The second challenge involves establishing effective training mechanisms using generated pseudo labels. 
Current approaches typically employ binary rewards \cite{dou2024integrating,zuo2025ttrl} that only supervise exact matches or entropy minimization \cite{agarwal2025unreasonable,wang2025test} that sharpens probability distributions. 
Yet such systems inadequately evaluate responses by failing to capture semantic similarities and partial correctness prevalent in open-ended medical answers, resulting in loss of valuable learning signals and suboptimal model convergence. 
These issues motivate us to design a bootstrapped TTT framework, which we term self-evolution, where the model converts its test-time responses into supervision signals and iteratively improves through a closed loop.

To overcome the limitations in existing approaches, we propose, to the best of our knowledge, \textit{the first \textbf{med}ical MLLM test-time self-\textbf{evo}lution framework} named \textbf{Med-Evo}, which addresses the limitation that previous works that focus solely on training data while neglecting the potential of leveraging test data for improving MLLMs.
A desirable self-evolution method should produce reliable pseudo targets and informative rewards, enabling stable label-free optimization as shown in Fig. \ref{Fig:teaser}(c).
Our framework consists of two key innovations. 
First, to address the inadequacy of majority voting mechanisms in generating reliable pseudo labels when medical VQA sampling produces divergent responses, we propose the Feature-based Pseudo Labeling (FPL) methodology, which utilizes semantic feature extraction to establish a centroid across all rollout responses and designates the nearest response as the pseudo label.
Second, recognizing that binary reward systems inadequately assess responses by only rewarding exact matches while potentially neglecting semantically similar correct answers, we propose the Hard-Soft Reward (HSR) approach, which combines binary rewards with semantic Jaccard similarity and semantic-derived feature distances to establish a more fine-grained scoring mechanism for medical VQA evaluation.
These innovations enable our framework to achieve superior medical MLLM adaptation performance without requiring additional labeled data.
We evaluate Med-Evo across multiple medical VQA datasets with different MLLM base models, including both general-domain and medical-specific fine-tuned MLLMs.
The experimental results indicate the effectiveness of our method compared to existing methods and establish its practical applicability in real clinical scenarios.

\section{Method}
\begin{figure}[t]
\centering
\includegraphics[width=\linewidth]{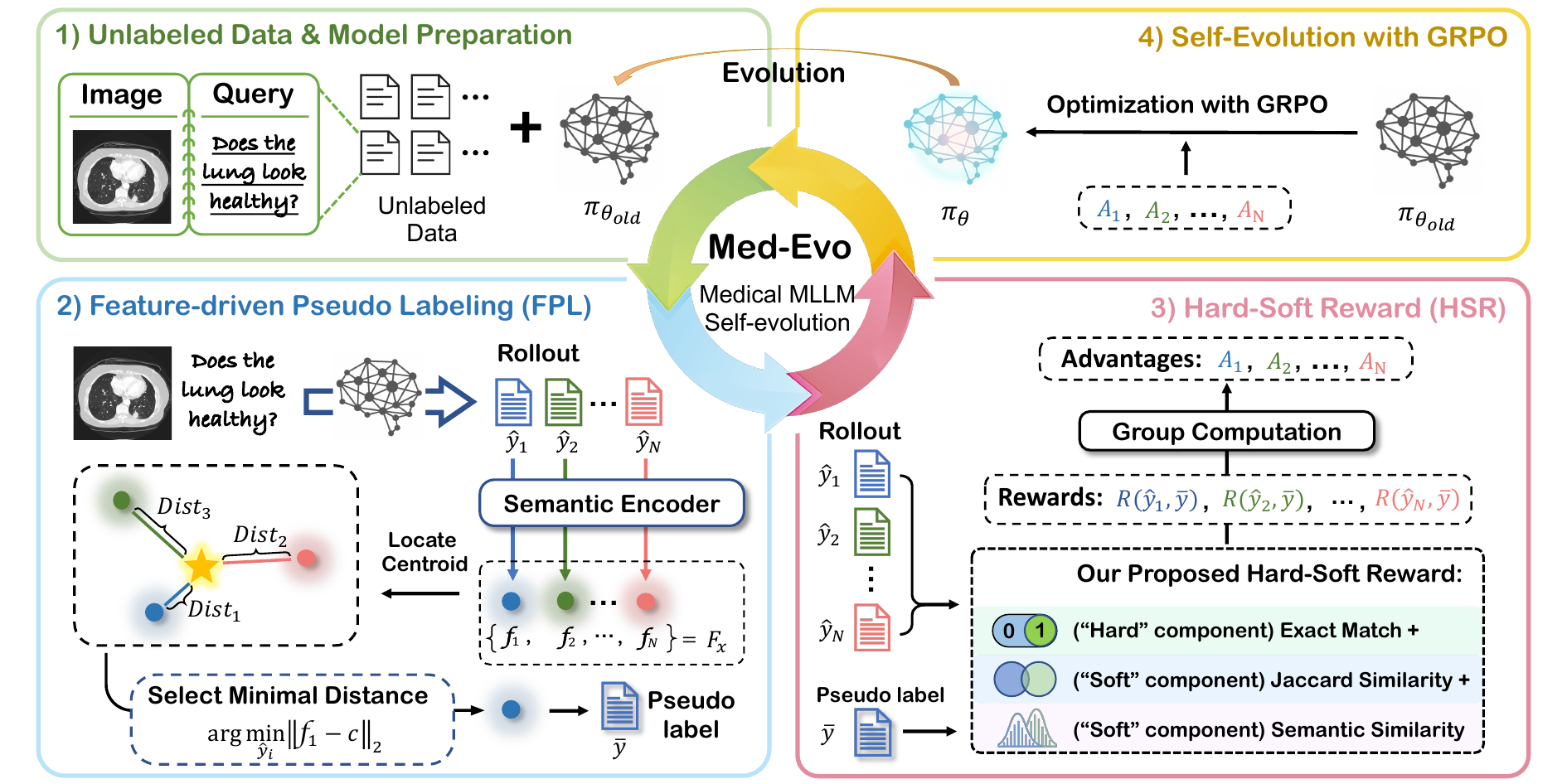}
\caption{Overview of our proposed  Med-Evo framework for test-time self-evolution with four stages: $1)$ we prepare unlabeled data and the policy model $\pi_{\theta_{old}}$; $2)$ we employ Feature-driven Pseudo Labeling (FPL) to generate reliable supervision signals from unlabeled data (Sec. \ref{FPL}); $3)$ we utilize the Hard-soft Reward (HSR) mechanism to provide hierarchical reward signals by incorporating finer-grained evaluation strategies (Sec. \ref{SHR}); $4)$ we leverage Group Relative Policy Optimization (GRPO) to achieve iterative unsupervised self-evolution through relative advantage computation (Sec. \ref{GRPO}).}
\label{Fig:main}
\end{figure}

Our proposed Med-Evo framework seeks to enable medical multimodal large language models (MLLMs) to achieve continuous performance optimization through unsupervised test-time self-evolution.
As illustrated in Fig. \ref{Fig:main}, each evolution cycle consists of four fundamental steps:
1) data preparation and model initialization for unlabeled test instances and parameter configuration;
2) parallel rollout for candidate responses generation and pseudo label synthesis using Feature-driven Pseudo Labeling (FPL) for each test case;
3) reward quantification and advantage estimation using our proposed Hard-soft Reward (HSR) mechanism;
and 4) iterative policy refinement based on obtained advantage scores.

\subsection{Feature-driven Pseudo Labeling}
\label{FPL}
Our Feature-driven Pseudo Labeling (FPL) mechanism addresses the challenge of generating reliable supervision signals from unlabeled test data. 
Instead of employing conventional majority voting approaches, FPL utilizes semantic coherence to establish robust consensus while mitigating the challenge of lexical heterogeneity inherent of generated responses in medical VQA tasks.

\noindent\textbf{Rollout Generation}. 
Given a test instance $x$ consisting of a medical image paired with a textual query, we generate a response rollout consists of $N$ distinct response candidates $\{\hat{y}_1, \hat{y}_2, \ldots, \hat{y}_N\}$ through stochastic sampling from the current policy distribution $\pi_{\theta_{old}}(\cdot | x)$. 
Each response generation within the rollout is conducted independently and stochastically, ensuring no cross-influence among the candidates.
To capture the semantic essence of each response beyond surface-level textual variations, we employ a semantic encoder $\mathbf{E}$ to extract high-dimensional feature representations, constructing the rollout of embeddings:

\begin{equation}
       F_x=\{f_1, f_2, \ldots, f_N\}, \quad f_i=\mathbf{E}(\hat{y}_i),
\label{eq:bert}
\end{equation}
where $F_x$ represents the feature set for all responses generated from instance $x$.

\noindent\textbf{Pseudo Label Selection.}
Based on these semantic embeddings, we can perform clustering analysis in the high-dimensional feature space rather than relying on strict string matching. 
Specifically, we locate the semantic centroid by averaging all embeddings $c=\frac{1}{N}\sum_{i=1}^{N}f_i$, where$f_i \in F_x.$
The pseudo label $\bar{y}$ for this case is then determined by identifying the candidate response whose semantic embedding exhibits the minimal distance to the computed semantic centroid:

\begin{equation}
\bar{y} = \arg\min_{\hat{y}_i} ||f_i - c||_2, \quad f_i \in F_x.
\label{eq:pseudo_label}
\end{equation}
This Feature-driven Pseudo Labeling strategy effectively provides a robust supervision signal generation in unsupervised test-time training, providing the foundation for further reward quantification and advantage estimation processes.

\subsection{Hard-soft Reward}
\label{SHR}
Traditional binary reward \cite{dou2024integrating,zuo2025ttrl} only encourages exact matches with reference labels while disregarding semantically equivalent answers through diverse lexical representations.
Our Hard-soft Reward (HSR) mechanism addresses this evaluation inadequacy by implementing adaptive reward computation strategies that provide finer-grained performance feedback for effective policy optimization.

\noindent\textbf{Binary Reward.}
We first establish the binary reward component $r_{\text{binary}}$ as the precision anchor of our reward system, ensuring stable optimization for responses with exact lexical correspondence. 
Specifically, this yields $r_{\text{binary}}(\hat{y}_i, \bar{y}) = 1$ when $\hat{y}_i = \bar{y}$ (indicating perfect matches) and $r_{\text{binary}}(\hat{y}_i, \bar{y}) = 0$ otherwise.

\noindent\textbf{Jaccard Similarity.}
While the strict binary matching ensures precision for exact matches, it fails to recognize partial correctness when responses contain overlapping terminology but differ in overall formulation. 
To address this limitation, we incorporate token-level assessment through Jaccard similarity:

\begin{equation}
r_{\text{jaccard}}(\hat{y}_i, \bar{y}) = \frac{|T(\hat{y}_i) \cap T(\bar{y})|}{|T(\hat{y}_i) \cup T(\bar{y})|}
\label{eq:jaccard_reward},
\end{equation}
where $T(\cdot)$ denotes the set of tokens in a response. 
This metric provides refined supervision signals that transcend the limitations of binary reward evaluation.

\noindent\textbf{Encoder-derived Semantic Similarity.}
While the aforementioned components address exact matching and token-level overlap, they fail to capture semantic equivalence in responses through diverse medical terminologies.
To quantify high-level semantic proximity, we employ contextualized embeddings to enable recognition of semantically equivalent expressions, which is formulated as:

\begin{equation}
r_{\text{semantic}}(\hat{y}_i, \bar{y}) = 1 - \frac{||\mathbf{E}(\hat{y}_i) - \mathbf{E}(\bar{y})||_2}{\max_j ||\mathbf{E}(\hat{y}_j) - \mathbf{E}(\bar{y})||_2}.
\label{eq:semantic_reward}
\end{equation}
This normalization bounds the reward to $[0,1]$ and rescales semantic distances relative to the rollout spread, producing more discriminative signals when candidates are diverse and more conservative ones when candidates are near-duplicates.

\noindent\textbf{Hard-soft Reward.}
Finally, we can construct our unified hard-soft reward as:
\begin{equation}
r_{\text{ours}} = \underbrace{ \alpha \cdot r_{\text{binary}}}_{\text{``hard" component}} + \underbrace{ \beta \cdot r_{\text{jaccard}} + (1-\alpha - \beta) \cdot r_{\text{semantic}}}_{\text{``soft" component}},
\label{eq:composite_reward}
\end{equation}
where $\alpha$ and $\beta$ are hyper-parameters that balance the contribution of each reward. 
This composite framework provides hierarchical evaluation signals, ranging from perfect matches through lexical overlap to semantic correspondence.
To accommodate diverse response types, we design an adaptive reward mechanism.
For closed-ended responses (e.g., Yes-No), we employ the binary reward solely. 
Conversely, for open-ended responses that include multiple expressions, we utilize the comprehensive hard-soft reward to capture nuanced correctness.

\subsection{Self-evolution with GRPO}
\label{GRPO}

After computing the rewards for each response in the same rollout, we employ Group Relative Policy Optimization (GRPO) \cite{shao2024deepseekmath} to optimize the model. 
GRPO provides stable policy updates by computing relative advantages within each rollout. 
For each reward \(r_i\) in the rollout, we compute the advantage as: 
\begin{equation}
A_i = \frac{r_i - \mu_r}{\sigma_r},\text{where} \quad \mu_r = \frac{1}{N} \sum_{j=1}^{N} r_j, \quad \sigma_r = \sqrt{\frac{1}{N} \sum_{j=1}^{N} (r_j - \mu_r)^2}.
\label{eq:bert}
\end{equation}

\noindent The model parameters \(\theta\) are updated using the policy gradient: 
\begin{equation}
\mathcal{J}_{\text{GRPO}}(\theta) = \mathbb{E}\left[\bar{\ell}(\theta) - \gamma D_{\mathrm{KL}}(\pi_\theta | \pi_{\theta_{old}})\right],
\label{eq:grpo_obj}
\end{equation}
where the clipped surrogate loss $\bar{\ell}(\theta)$ is computed as:
\begin{equation}
\bar{\ell}(\theta) = \frac{1}{N}\sum_{i=1}^{N} \min\left(\rho_i(\theta) A_i, \text{clip}_\varepsilon(r_i(\theta)) A_i\right),
\label{eq:clipped_loss}
\end{equation}
with $\rho_i(\theta) = \frac{\pi_\theta(\hat{y}_i|x)}{\pi_{\theta_{old}}(\hat{y}_i|x)}$ representing the probability ratio between current and reference policy models. 
The clipping function constrains ratios within $[1-\varepsilon, 1+\varepsilon]$ to prevent excessive updates, while the KL divergence term with coefficient $\gamma$ regularizes optimization for stable convergence.
After each optimization iteration, the reference policy model $\pi_{\theta_{old}}$ is updated with the current optimized model $\pi_{\theta}$, enabling the model to achieve progressive unsupervised self-evolution.

\section{Experiments}
\noindent\textbf{Dataset.} 
We use only test data (without label) to update MLLMs with our self-evolution approach based on three English-based medical VQA tasks:
1) SLAKE \cite{liu2021slake} with 642 images and over 7,000 QA pairs; 
2) VQA-Rad \cite{lau2018dataset} with 315 images and 3,515 QA pairs; 
3) VQA-Med \cite{ben2019vqa} with 4,200 images and 15,292 QA pairs.

\begin{table}[b]
\centering
\fontsize{8pt}{8pt}\selectfont
\addtocounter{table}{0}
\renewcommand{\arraystretch}{1.35}
\setlength{\tabcolsep}{1.2mm}
{
\caption{Performance comparison on three medical VQA datasets using Qwen2.5-VL-3B-Instruct as the base model. \textbf{Bold} indicates the best performance for each metric.}
\label{tab:qwen}
\begin{tabular}{c|ccc|ccc|ccc}
\hline
\hline
 & \multicolumn{3}{c|}{SLAKE} & \multicolumn{3}{c|}{VQA-Rad} & \multicolumn{3}{c}{VQA-Med} \\
\hline
 Method & Acc & Recall & Rouge & Acc & Recall & Rouge & Acc & Recall & Rouge \\

\hline
Base Model & 68.73 & 34.70 & 43.97 & 68.53 & 24.00 & 32.03 & 56.88 & 12.57 & 27.45 \\
EN-INF \cite{agarwal2025unreasonable} & 65.92 & 35.55 & 45.23 & 67.73 & 24.00 & 32.42 & 55.96 & 10.18 & 25.16\\
TTRV \cite{singh2025ttrv} & 63.66 & 33.00 & 43.48 & 62.55 & 19.50 & 29.68 & 55.04 & 10.78 & 25.75\\
TTRL \cite{zuo2025ttrl} & 72.68 & 35.41 & 48.00 & 68.13 & 24.00 & 33.17 & 55.96 & 9.58 & 25.28\\
Ours & \textbf{78.87} & \textbf{39.38} & \textbf{51.01} &  \textbf{69.32} & \textbf{25.00} & \textbf{33.72} & \textbf{57.79} & \textbf{14.97} & \textbf{27.96} \\
\hline
\hline
\end{tabular}}
\end{table}

\noindent\textbf{Implementation Details.} 
We use Qwen2.5-VL-3B-Instruct \cite{bai2025qwen2} and MedVLM-R1 \cite{pan2025medvlm} as our base model for validating the broad applicability of our method on both general-purpose and medical-specialized MLLMs.
We utilize Robert-v1 \cite{liu2019roberta} as semantic encoder.
The learning rate is set to 5$\times10^{-7}$ with batch size 4.
The temperature is set to 0.6 and top-$p$ value is 0.95.
In each rollout, we generate 32 responses for pseudo label selection and use the first 16 for GRPO optimization.
We set the $\alpha$ to 0.85 and $\beta$ to 0.05 in Eq. \ref{eq:composite_reward}.
All experiments were carried out on 4 NVIDIA RTX A6000 GPUs. 
We evaluate model performance using Accuracy on closed-ended questions and ROUGE-1 and Recall on open-ended questions.

\noindent\textbf{Comparison with State-of-the-Arts.} 
We compared our Med-Evo framework with several state-of-the-art test-time training methods: 1) \textbf{EN-INF} \cite{agarwal2025unreasonable} is developed for LLM by minimizing the entropy during the inference period; 2) \textbf{TTRV} \cite{singh2025ttrv} designs unsupervised reward signals based on frequency and entropy; 3) \textbf{TTRL} \cite{zuo2025ttrl} proposes test-time training based on majority voting and reinforcement learning.
To clearly demonstrate the improvements brought by test-time training, we also include the \textbf{Base Model} results in our comparison, which directly performs inference on the datasets without any test-time adaptation.

\noindent\textbf{Results on Qwen2.5-VL-3B-Instruct Across Three Datasets.} 
Table \ref{tab:medvlm} demonstrates the superior performance of our Med-Evo framework compared to baseline methods using Qwen2.5-VL-3B-Instruct as the base model. 
Across all three medical VQA datasets, our approach consistently achieves the highest performance.
On SLAKE, our method achieves the accuracy of 78.87\% on close-ended questions, demonstrating a substantial 10.14\% improvement over the base model.
Notably, for open-ended questions, our framework excels with 39.38\% recall and 51.01\% ROUGE scores, outperforming the second-best method by 3.97\% and 3.01\% respectively. 
The VQA-Rad results further corroborate our method's effectiveness, with our approach attaining 69.32\% accuracy, 25.00\% recall, and 33.72\% ROUGE score, consistently surpassing all competing methods. 
On VQA-Med, our framework achieves satisfied performance with 57.79\% accuracy, while maintaining competitive recall and ROUGE scores.  
These comprehensive results validate the effectiveness of our MLLM self-evolution approach for medical VQA.

\begin{table}[t]
\centering
\fontsize{8pt}{8pt}\selectfont
\addtocounter{table}{0}
\renewcommand{\arraystretch}{1.35}
\setlength{\tabcolsep}{1.2mm}
{
\caption{Performance comparison on three medical VQA datasets using  MedVLM-R1 as the base model. \textbf{Bold} indicates the best performance for each metric.}
\label{tab:medvlm}
\begin{tabular}{c|ccc|ccc|ccc}
\hline
\hline
 & \multicolumn{3}{c|}{SLAKE} & \multicolumn{3}{c|}{VQA-Rad} & \multicolumn{3}{c}{VQA-Med} \\
\hline
 Method & Acc & Recall & Rouge & Acc & Recall & Rouge & Acc & Recall & Rouge \\

\hline
Base Model & 61.41 & 17.99 & 26.82 & 49.00 & 13.00 & 19.33 & 44.04 & 9.58 & 23.02 \\
EN-INF \cite{agarwal2025unreasonable} & 68.73 & 25.21 & 33.50  & 58.57 & 17.00 & 24.58 & 46.79 & 10.18 & 23.02 \\
TTRV \cite{singh2025ttrv} & 68.45 & 25.64 & 33.74 & 59.36 & 17.50 & 25.02 & 44.95 & 10.18 & 21.83\\
TTRL \cite{zuo2025ttrl} &  71.27 & 26.20 & 36.72 & 57.37 & 17.00 & 25.08 & 38.53 & 6.59 & 20.23\\
Ours & \textbf{71.55} & \textbf{30.03} & \textbf{39.66} & \textbf{61.75} & \textbf{18.00} & \textbf{25.37} & \textbf{48.62} & \textbf{10.18} & \textbf{23.42}\\
\hline
\hline
\end{tabular}}
\end{table}

\noindent\textbf{Results on MedVLM-R1 Across Three Datasets.}
Table \ref{tab:medvlm} presents the experimental results using MedVLM-R1 as the base model, further demonstrating the generalizability of our approach across different foundation architectures.
On the SLAKE dataset, our approach achieves the highest accuracy of 71.55\% for closed-ended questions among all competing methods, while also exhibiting remarkable effectiveness on open-ended questions with 30.03\% recall and 39.66\% ROUGE scores, outperforming the second-best method by 3.83\% and 2.94\% respectively.
On both VQA-Rad and VQA-Med datasets, our method demonstrates substantial improvements over the base model, with notable gains of 12.75\% and 4.58\% in accuracy respectively, while consistently outperforming all competing methods across both closed-ended and open-ended question evaluations.
These results, combined with the findings from Table \ref{tab:qwen}, indicate that our self-evolution approach maintains stable performance improvements regardless of the base model's initial capabilities, confirming the robustness and effectiveness of our method across both general-purpose and medical-specific MLLMs.

\begin{figure}[t]
\centering
\includegraphics[width=\linewidth]{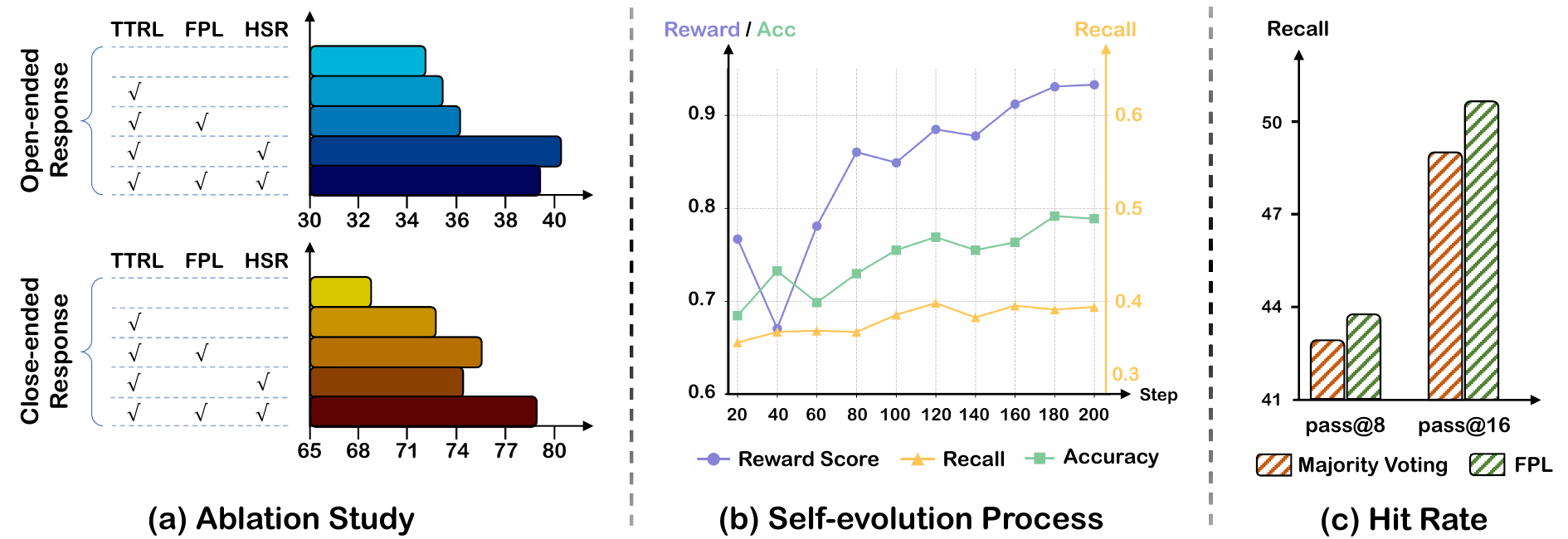}
\caption{ Illustrative experiments based on Qwen2.5-VL-3B-Instruct on SLAKE dataset. (a) ablation study; (b) averaged reward score and model performance during the whole evolution process; (c) the hit rate comparison using base model.}
\label{Fig:ablation}
\end{figure}

\noindent\textbf{Ablation Study.} 
To validate the effectiveness of each component in our framework, we conducted ablation studies as shown in Fig. \ref{Fig:ablation}(a). 
We evaluated five configurations: 
1) direct inference without self-evolution; 
2) basic test-time reinforcement learning; 
3) test-time reinforcement learning with feature-driven pseudo labeling; 
4) test-time reinforcement learning incorporating hard-soft reward; and 
5) the complete framework integrating all components. 
The results indicate that each component contributes meaningful performance improvements.

\noindent\textbf{Self-evolution process.}
Figure \ref{Fig:ablation}(b) shows the self-evolution progress. 
The purple line represents the reward score using exponential moving average across the whole process, while the green and yellow lines show the accuracy and recall on closed-ended and open-ended questions, respectively. 
The results demonstrate a clear positive correlation between reward score and model performance, confirming that our hard-soft Reward effectively guides the self-evolution process.

\noindent\textbf{Hit rate.}
Figure \ref{Fig:ablation}(c) compares the hit rate performance between Majority Voting and our proposed FPL method across different response sampling numbers (i.e., pass@8 and pass@16).
Hit rate is the proportion of cases where the pseudo label matches the ground truth.
The orange bars denote Majority Voting, while the green bars denote FPL. 
The orange bars represent the hit rate achieved by Majority Voting, while the green bars show the performance of FPL method.
The results demonstrate that FPL consistently outperforms Majority Voting in both settings, validating the effectiveness of our feature-driven pseudo labeling.

\section{Conclusion}
This paper proposes the Med-Evo for stable MLLM test-time self-evolution without requiring extensive labeled data, which incorporates Feature-driven Pseudo Labeling (FPL) for generation supervision signals and Hard-Soft Reward (HSR) for fine-grained performance feedback.
Comprehensive experiments across multiple medical VQA tasks with different base models demonstrate the effectiveness and generalizability of our approach. 
The broader impact of our work lies in establishing a practical framework that enables medical MLLMs to achieve performance improvements without requiring labeled data, potentially facilitating AI-assisted healthcare solutions in resource-constrained clinical environments.

\bibliographystyle{splncs04}
\bibliography{reference}

\begin{thebibliography}{10}
\providecommand{\url}[1]{\texttt{#1}}
\providecommand{\urlprefix}{URL }
\providecommand{\doi}[1]{https://doi.org/#1}

\bibitem{abdullahi2024learning}
Abdullahi, T., Singh, R., Eickhoff, C.: Learning to make rare and complex diagnoses with generative ai assistance: qualitative study of popular large language models. JMIR Medical Education  \textbf{10}(1),  e51391 (2024)

\bibitem{agarwal2025unreasonable}
Agarwal, S., Zhang, Z., Yuan, L., Han, J., Peng, H.: The unreasonable effectiveness of entropy minimization in llm reasoning. arXiv preprint arXiv:2505.15134  (2025)

\bibitem{bai2025qwen2}
Bai, S., Chen, K., Liu, X., Wang, J., Ge, W., Song, S., Dang, K., Wang, P., Wang, S., Tang, J., et~al.: Qwen2. 5-vl technical report. arXiv preprint arXiv:2502.13923  (2025)

\bibitem{ben2019vqa}
Ben~Abacha, A., Hasan, S.A., Datla, V.V., Demner-Fushman, D., M{\"u}ller, H.: Vqa-med: Overview of the medical visual question answering task at imageclef 2019. In: Proceedings of CLEF (Conference and Labs of the Evaluation Forum) 2019 Working Notes. 9-12 September 2019 (2019)

\bibitem{dou2024integrating}
Dou, C., Zhang, Y., Jin, Z., Jiao, W., Zhao, H., Zhao, Y., Tao, Z.: Integrating physician diagnostic logic into large language models: Preference learning from process feedback. In: Findings of the Association for Computational Linguistics: ACL 2024. pp. 2453--2473 (2024)

\bibitem{goel2023llms}
Goel, A., Gueta, A., Gilon, O., Liu, C., Erell, S., Nguyen, L.H., Hao, X., Jaber, B., Reddy, S., Kartha, R., et~al.: Llms accelerate annotation for medical information extraction. In: machine learning for health (ML4H). pp. 82--100. PMLR (2023)

\bibitem{he2024pefomed}
He, J., Li, P., Liu, G., He, G., Chen, Z., Zhong, S.: Pefomed: Parameter efficient fine-tuning of multimodal large language models for medical imaging. arXiv preprint arXiv:2401.02797  (2024)

\bibitem{hosseini2025cg}
Hosseini, P., Bohdal, O., Ceritli, T., Castro, I., Purver, M., Ozay, M., Michieli, U.: Cg-ttrl: Context-guided test-time reinforcement learning for on-device large language models. arXiv preprint arXiv:2511.06430  (2025)

\bibitem{lau2018dataset}
Lau, J.J., Gayen, S., Ben~Abacha, A., Demner-Fushman, D.: A dataset of clinically generated visual questions and answers about radiology images. Scientific data  \textbf{5}(1),  180251 (2018)

\bibitem{liu2021slake}
Liu, B., Zhan, L.M., Xu, L., Ma, L., Yang, Y., Wu, X.M.: Slake: A semantically-labeled knowledge-enhanced dataset for medical visual question answering. In: 2021 IEEE 18th international symposium on biomedical imaging (ISBI). pp. 1650--1654. IEEE (2021)

\bibitem{liu2019roberta}
Liu, Y., Ott, M., Goyal, N., Du, J., Joshi, M., Chen, D., Levy, O., Lewis, M., Zettlemoyer, L., Stoyanov, V.: Roberta: A robustly optimized bert pretraining approach. arXiv preprint arXiv:1907.11692  (2019)

\bibitem{liu2025infimed}
Liu, Z., Hou, Z., Zhu, G., Sang, Z., Xie, C., Yang, H.: Infimed: Low-resource medical mllms with advancing understanding and reasoning. arXiv preprint arXiv:2505.23867  (2025)

\bibitem{ma2025medla}
Ma, S., Huang, J., Yang, B., Zhang, F., Wu, J., Shen, Y., Fan, G., Zhang, Z., Zang, Z.: Medla: A logic-driven multi-agent framework for complex medical reasoning with large language models. arXiv preprint arXiv:2509.23725  (2025)

\bibitem{pan2025medvlm}
Pan, J., Liu, C., Wu, J., Liu, F., Zhu, J., Li, H.B., Chen, C., Ouyang, C., Rueckert, D.: Medvlm-r1: Incentivizing medical reasoning capability of vision-language models (vlms) via reinforcement learning. In: International Conference on Medical Image Computing and Computer-Assisted Intervention. pp. 337--347. Springer (2025)

\bibitem{rawat2024diversitymedqa}
Rawat, R., McBride, H., Ghosh, R., Nirmal, D., Moon, J., Alamuri, D., Brien, S.O., Zhu, K.: Diversitymedqa: A benchmark for assessing demographic biases in medical diagnosis using large language models. In: Proceedings of the Third Workshop on NLP for Positive Impact. pp. 334--348 (2024)

\bibitem{shao2024deepseekmath}
Shao, Z., Wang, P., Zhu, Q., Xu, R., Song, J., Bi, X., Zhang, H., Zhang, M., Li, Y., Wu, Y., et~al.: Deepseekmath: Pushing the limits of mathematical reasoning in open language models. arXiv preprint arXiv:2402.03300  (2024)

\bibitem{singh2025ttrv}
Singh, A., Marjit, S., Lin, W., Gavrikov, P., Yeung-Levy, S., Kuehne, H., Feris, R., Doveh, S., Glass, J., Mirza, M.J.: Ttrv: Test-time reinforcement learning for vision language models. arXiv preprint arXiv:2510.06783  (2025)

\bibitem{sood2025medical}
Sood, A., Pattnaik, T., Malhotra, R., Nayyar, V., Narayan, B., Mishra, D., Surya, V., et~al.: Medical imaging privacy: a systematic scoping review of key parameters in dataset construction and data protection. Journal of Medical Imaging and Radiation Sciences  \textbf{56}(5),  101914 (2025)

\bibitem{sysoykova2025federated}
Sysoykova, E., Anzengruber-Tanase, B., Haslgrubler, M., Seidl, P., Ferscha, A.: Federated few-shot learning for epileptic seizure detection under privacy constraints. arXiv preprint arXiv:2512.13717  (2025)

\bibitem{tajbakhsh2021guest}
Tajbakhsh, N., Roth, H., Terzopoulos, D., Liang, J.: Guest editorial annotation-efficient deep learning: the holy grail of medical imaging. IEEE transactions on medical imaging  \textbf{40}(10),  2526--2533 (2021)

\bibitem{wac2025capturing}
Wac, M., Santos-Rodriguez, R., McWilliams, C., Bourdeaux, C.: Capturing requirements for a data annotation tool for intensive care: Experimental user-centered design study. JMIR Human Factors  \textbf{12}(1),  e56880 (2025)

\bibitem{wang2025test}
Wang, H., Yu, Y., Zheng, H., Zhang, T.: Test-time adaptation of medical vision-language models with mixture of modality experts. In: Proceedings of the 33rd ACM International Conference on Multimedia. pp. 4649--4658 (2025)

\bibitem{wang2025hicur}
Wang, Z., Fang, M., Tang, L., Tian, J., Dong, D.: Hicur-npc: Hierarchical feature fusion curriculum learning for multi-modal foundation model in nasopharyngeal carcinoma. IEEE Transactions on Medical Imaging  (2025)

\bibitem{wei2025first}
Wei, L., Li, Y., Wang, C., Wang, Y., Kong, L., Huang, W., Sun, L.: First sft, second rl, third upt: Continual improving multi-modal llm reasoning via unsupervised post-training. arXiv preprint arXiv:2505.22453  (2025)

\bibitem{xu2025perceive}
XU, D., Yang, X., Li, Y., Miao, J., Li, J., Heng, P.A.: Perceive and calibrate: Analyzing and enhancing robustness of medical multi-modal large language models. arXiv preprint arXiv:2512.21964  (2025)

\bibitem{xu2024towards}
Xu, S., Zhou, Y., Liu, Z., Wu, Z., Zhong, T., Zhao, H., Li, Y., Jiang, H., Pan, Y., Chen, J., et~al.: Towards next-generation medical agent: How o1 is reshaping decision-making in medical scenarios. arXiv preprint arXiv:2411.14461  (2024)

\bibitem{zhang2025aqa}
Zhang, H., Guo, J., Iwasawa, Y., Matsuo, Y.: Aqa-ttrl: Self-adaptation in audio question answering with test-time reinforcement learning. arXiv preprint arXiv:2510.05478  (2025)

\bibitem{zhang2024potential}
Zhang, Y., Pan, Y., Zhong, T., Dong, P., Xie, K., Liu, Y., Jiang, H., Wu, Z., Liu, Z., Zhao, W., et~al.: Potential of multimodal large language models for data mining of medical images and free-text reports. Meta-Radiology  \textbf{2}(4),  100103 (2024)

\bibitem{zhi2025medgr}
Zhi, W., Guo, J., Li, S.: Medgr$^2$: Breaking the data barrier for medical reasoning via generative reward learning. arXiv preprint arXiv:2508.20549  (2025)

\bibitem{zhu2025toward}
Zhu, W., Dong, X., Li, X., Qiu, P., Chen, X., Razi, A., Sotiras, A., Su, Y., Wang, Y.: Toward effective reinforcement learning fine-tuning for medical vqa in vision-language models. arXiv preprint arXiv:2505.13973  (2025)

\bibitem{zuo2025ttrl}
Zuo, Y., Zhang, K., Sheng, L., Qu, S., Cui, G., Zhu, X., Li, H., Zhang, Y., Long, X., Hua, E., et~al.: Ttrl: Test-time reinforcement learning. arXiv preprint arXiv:2504.16084  (2025)

\end{thebibliography}

\end{document}